# Content-Based Smart E-Mail Dispatcher Using Large Language Models

K. Paramesha(✉), K R Sriram, Sujan Shetty, Shamanth Kishore, and R. Tejaswini

Vidyavardhaka College of Engineering, Mysuru, India
paramesha.k@vvce.ac.in
https://vvce.ac.in/departments/department-of-computer-science-engineering/

**Abstract.** Email communication has become an integral part of personal and professional life, but handling its vast volume is still a significant issue for large organizations. Manual perusal of emails and forwarding their contents and attachments to intended recipients using other instant messaging platforms has proven to be error-prone and time-consuming leading to losses in terms of productivity and creating undue stress. The main objective of this paper is to explore an alternative mechanism that is to automate the task of dispatching emails based on their contents to the respective WhatsApp groups of students of various semesters of programs in an engineering college, facilitating a smooth flow of information from one end to another end in an organization. The dispatcher system is built using agents querying large language models (LLMs) to enable it to analyze the contents of emails and route them to the relevant groups of students for their information and consumption. The system harnesses the capabilities of LLMs in analyzing the textual contents for decision-making. With a well-structured agent framework prompting that includes email content as input with instructions and context, the system figures out the relevant groups to which the email message is dispatched thus providing the required information on time. The proposed system does not rely on labeled datasets and provides several benefits, including enhanced productivity and a reduction in the cognitive load associated with reading emails.

## 1 Introduction

Emails have gone from obscure to ubiquitous in half a century. Each day, on average, a person can deal with as many as 21 to 26 emails in a short span of time, creating an overwhelming cognitive load. Studies highlight that each email interruption costs 64 s to recover focus, significantly affecting productivity [1]. With over 4.4 billion email accounts globally and more than 320 billion emails exchanged daily, manual peruse remains inefficient and stressful for employees, as

evidenced by increased stress levels and reduced productivity when emails are checked continuously [7]. Therefore, decreasing the usage of emails might not be the right solution. Although continuous interruption of work due to emails can have negative impacts, through proper techniques. These findings highlight the need for an automated solution to reduce the cognitive burden, and time consumption to effectively manage emails [11] caused by manual email perusal.

In this paper, we study the problems associated with manual reading and forwarding email messages received in the department mailbox to the concerned students for their information and follow-up actions. From the students' point of view, these academic notifications and circulars are important to keep in sync with the academic calendar of events. Figure 1 shows the structure of the information flow from higher authorities to students through the respective department. The departments receive emails containing different types of information from higher-ups. These email messages and attachments are routed to the students registered to the respective batch's Whatsapp group. The assigned staff in the department is responsible for manually going through the emails and forwarding them to the respective WhatsApp group for student consumption. This process is burdening the staff, affecting their productivity. Operating manually is bound to face a lot of issues. The solution to these types of manual operations is automation. With the latest advancements in natural language processing (NLP) techniques, checking emails and forwarding them to the intended recipients could be automated. The level of automation demands analyzing their contents to identify the target WhatsApp groups to which the email message is relevant rather than just channeling incoming emails to groups. As email messages are from diverse sources, understanding the sense of the content is key to providing an efficient solution. With efficient development and deployment of this type of system, two challenges are addressed. 1) Manual email checking and forwarding operations can be eliminated significantly thereby productivity of the staff is enhanced. 2) Since students frequently use instant messaging apps such as WhatsApp and Telegram, posting the latest information will enable them to stay updated with the latest happenings.

## 2 Related Work

The potential of LLMs is enormous and has demonstrated a variety of use cases on multi-modal data [3]. LLMs are becoming more relevant and useful when annotated datasets are unavailable as preparing them requires extensive manual effort. They can classify a URL link that is commonly seen embedded in email as malicious or not by one-shot prompting [10]. LLMs trained on large data have demonstrated the ability to understand input text and generate suitable responses as per the requirements of the applications such as generating documentation [2], machine translation [5], medical transcription [12], malware detection [9], etc. Many automated machine learning applications have been developed to analyze email content for classification tasks [6] and optimize workload management online [14]. There are some automated services online [15] wherein the system is configured to forward the received emails sent by a particular account as WhatsApp messages without content analysis. Keywords in the subject line can drive the email forwarding to WhatApp [8]. Artificial Intelligence(AI) is being integrated into platforms that use LLMS to perform a variety of automation tasks specifically in question and answering mode in chat and email applications. If the task is complex, it is decomposed into subtasks which solved through agents making the model adaptable and scalable [13].

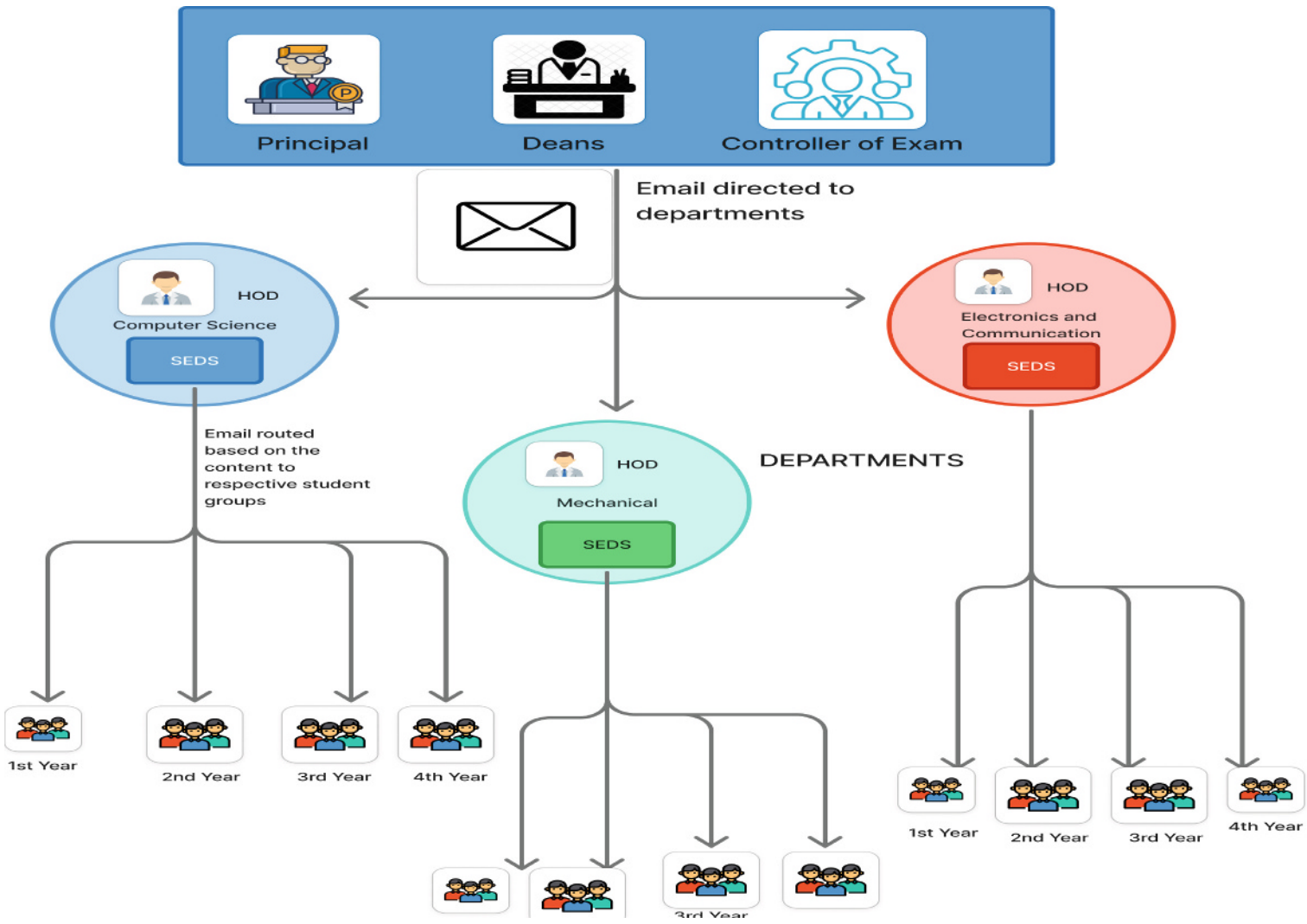

**Fig. 1.** Information flow from higher authorities to students.

# 3 Methodology

The proposed solution to the challenges is to build and validate an automated system before deployment. The system functionality is divided into three steps. Step 1: Data capturing, in this step reading emails is performed. In the development phase, the data in the textual format is obtained by reading email messages and their attachments offline, whereas, in the deployment phase, email messages and their attachments are read from the mail inbox. Step 2: Content Analysis, in which the content is analyzed by agents using LLM through prompts. Step 3: Dispatching emails, in which the email message is posted to WhatsApp groups specified by LLM. The important and core computation is the content processing step which depends on the LLM and design of the prompt.

## 3.1 Data Collection

The system is a data-driven model that analyzes email content to make decisions for dispatching them to appropriate WhatsApp groups. Therefore data collection and analysis to know its sources and categories are of primary importance to design prompt for LLMs. The dataset used for model validation comprises over 1000 carefully selected email messages related to students and faculty members.

### 3.2 AI Agent Prompt Engineering

Prompt Engineering refers to the techniques for designing prompts for LLMs to generate accurate, relevant, and contextually appropriate responses. We employ the Langchain framework to facilitate agents to perform the task step by step. Figure 2 shows a smart email dispatcher system(SEDS) to route the email to WhatsApp groups using custom tools used by the agents. Initially, the category of email content is identified by an agent and then the content is passed to the respective agent for the follow-up action. The agent for circular notifications and announcements needs to identify which of the target WhatsApp groups to which the message needs to be transferred. Having a WhatsApp group for each batch of students by their joined academic year and one faculty member group, the agents use the custom-built dispatcher to post email attachments and messages recommended by the LLM.

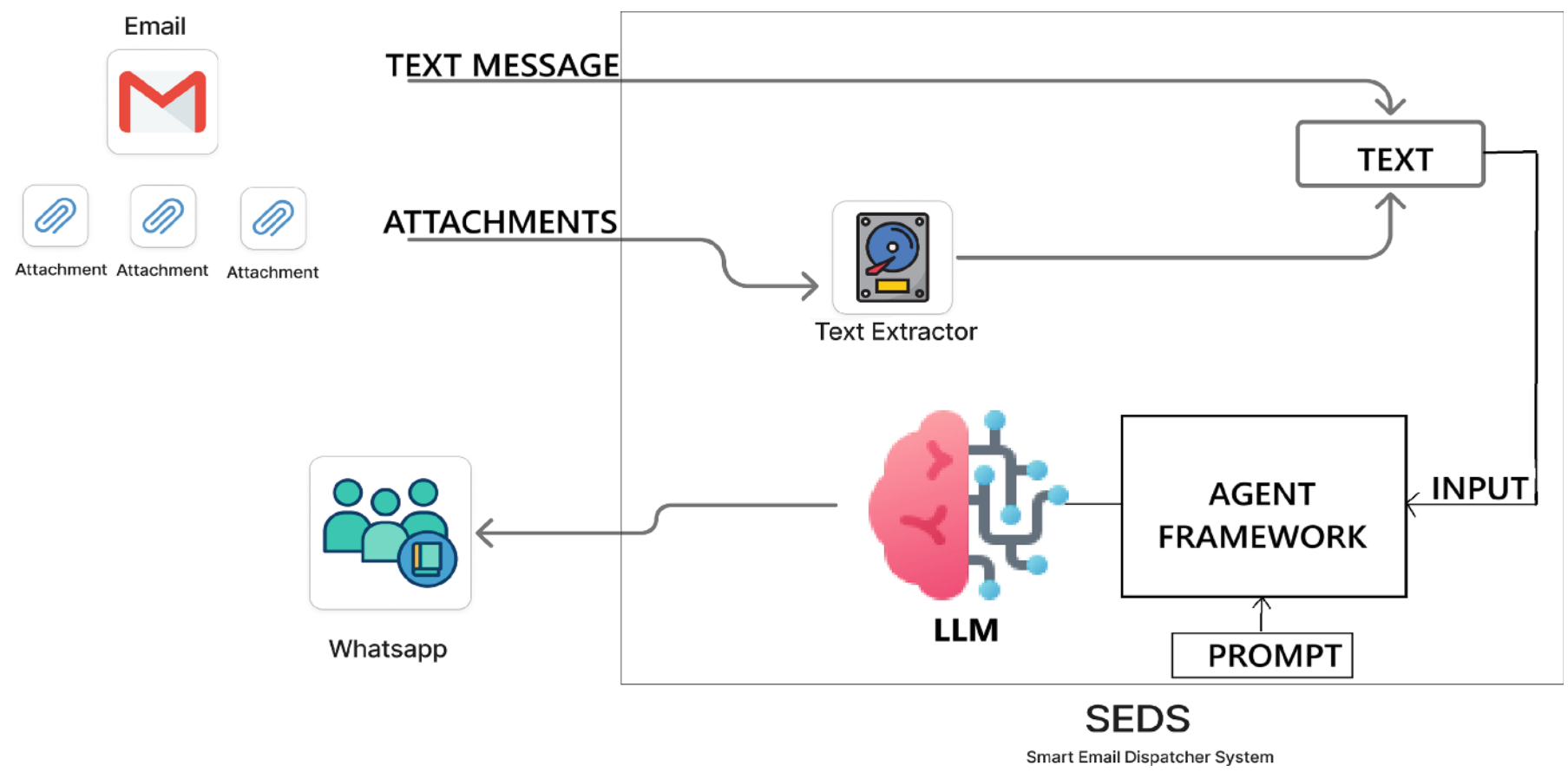

**Fig. 2.** Email dispatcher system using agent-based framework.

## 4 Results and Discussion

The results of the proposed model using various LLMs are furnished in Table 1. The agents query the LLM using Apis. We consider the perfect classification if the model identifies all the target WhatsApp groups. Even a single extra or missing target group will be considered a misclassification. ChatGPT and Gemini models perform much better than Deepseek and Perplexity AI models. When we closely observed misclassification results, we found that the model would recommend one or more groups other than targeted ones, especially in the case of circulars regarding valuation and revaluation, which are intended only for faculty members and are recommended to be shared with the student groups as well. Circulars mentioning only the batch series without year were able to map to the correct target groups.

**Table 1.** Accuracies of models on the dataset.

| MODELS/PLATFORMS | ACCURACY (in %) |
|---|---|
| Gemini | 93.5 |
| DeepSeek | 80 |
| Perplexity | 82.5 |
| ChatGPT | 92 |

### 4.1 Ablation of LLM

The replacement of LLM with a machine learning/deep learning model poses a lot of challenges for yielding the desired results. A labeled dataset is needed for training and validating the machine learning model to identify the target WhatsApp groups. Moreover, multiple selection of groups requires altogether a different technique in obtaining the expected results as machine learning prediction favours the best label. Clustering or any unsupervised technique will not lead to the solutions because the objectives are different. The WhatsApp group batches are not fixed as a new batch is added every year. However, if the system is implemented by overcoming all the challenges of the machine learning/deep learning models, the model's performance in the content analysis will not match that of LLMs as benchmarks have proven LLMs' supremacy in NLP tasks [4].

## 5 Conclusion and Future Work

The smart email dispatcher work is a giant leap forward in the automation and streamlining of email management for institutions and organizations. It simplifies workload management with the integration of advanced technologies such as transformer-based models and unsupervised learning, the system addresses challenges like semantic understanding, unstructured data processing, and dynamic email categorization. This results in a robust solution that can extract, classify, and route emails with remarkable accuracy and efficiency. Through automation, the system not only reduces the time and effort involved in handling manual emails but also minimizes errors and ensures consistency in performance. Its adaptability to changing communication patterns and the ability to process diverse formats like PDFs, images, and Word documents make it a versatile tool for modern workflows. This project's successful implementation showcases AI's potential to change the nature of operations. This is because it allows users to focus on high-priority tasks and strategic decisions, thereby improving productivity and efficiency in operations. This project is a stepping stone for future innovations in email management, which would eventually lead to smarter, adaptive communication systems across domains. In future work, we plan to provide email messages in a concise manner where the email messages will be summarized and pumped to WhatsApp channels for quick and clear notifications of events and information to the students' community.